\documentclass{article}% For LaTeX2e
\usepackage[preprint, nonatbib]{neurips_2024}
\usepackage[numbers]{natbib}

\usepackage{microtype}
\usepackage{hyperref}
\usepackage{url}
\usepackage{booktabs}

\usepackage{graphicx}
\usepackage{subcaption}
\usepackage{wrapfig}
\usepackage{tikz}

\usepackage{caption} 
\usepackage[most]{tcolorbox}
\newtcolorbox{highlight}[1][]{
    enhanced,
    colback=yellow!10,
    colframe=gray!30,
    boxrule=0.5pt,
    arc=2pt,
    leftrule=3pt,
    rightrule=3pt,
    toprule=1pt,
    bottomrule=1pt,
    breakable,
    #1
}

\makeatletter
\def\blfootnote{\xdef\@thefnmark{}\@footnotetext}
\makeatother

\title{From Artificial Needles to Real Haystacks: Improving Retrieval Capabilities in LLMs by Finetuning on Synthetic Data}

\author{Zheyang Xiong, Vasilis Papageorgiou, Kangwook Lee, Dimitris Papailiopoulos\\
\\
University of Wisconsin-Madison
}

\begin{document}

\maketitle

\begin{abstract}
Recent studies have shown that Large Language Models (LLMs) struggle to accurately retrieve information and maintain reasoning capabilities when processing long-context inputs. To address these limitations, we propose a finetuning approach utilizing a carefully designed synthetic dataset comprising numerical key-value retrieval tasks. Our experiments on models like GPT-3.5 Turbo and Mistral 7B demonstrate that finetuning LLMs on this dataset significantly improves LLMs' information retrieval and reasoning capabilities in longer-context settings. We present an analysis of the finetuned models, illustrating the transfer of skills from synthetic to real task evaluations (e.g., $10.5\%$ improvement on $20$ documents MDQA at position $10$ for GPT-3.5 Turbo). We also find that finetuned LLMs' performance on general benchmarks remains almost constant while LLMs finetuned on other baseline long-context augmentation data can encourage hallucination (e.g., on TriviaQA, Mistral 7B finetuned on our synthetic data cause no performance drop while other baseline data can cause a drop that ranges from $2.33\%$ to $6.19\%$). Our study highlights the potential of finetuning on synthetic data for improving the performance of LLMs on longer-context tasks.
\end{abstract}

\blfootnote{Email: \texttt{<zheyang@cs.wisc.edu>}. Correspondence: \texttt{<dimitris@papail.io>}.}
\blfootnote{Code available at: \texttt{github.com/edixiong/artificial-needles}}

\section{Introduction}
Recent studies have revealed that Large Language Models (LLMs) struggle to accurately retrieve information and maintain reasoning capabilities when processing longer context inputs or when retrieval is required across different parts of their context \citep{liu2023lost, levy2024same}. These limitations hinder their performance on tasks that involve processing and reasoning over extensive textual information, such as summarization or question answering over long passages. 

To address these challenges, we propose a novel approach that involves finetuning LLMs on a carefully designed fully numerical \emph{synthetic} dataset containing key-value dictionary retrieval tasks ({\it i.e.,} see Fig.~\ref{fig:simpledict_ntemplate} for an example of such a task). We conduct extensive experiments on popular LLMs, including GPT-3.5 Turbo \citep{chatgpt} and Mistral 7B \citep{jiang2023mistral}, and find that our method improves their performance on both information retrieval and long-context reasoning.

\par Specifically, our approach mitigates the ``lost-in-the-middle" phenomenon identified by \citet{liu2023lost} and significantly improves performance on the FLenQA benchmark \citep{levy2024same} that measures LLMs' long-context reasoning capability. Interestingly, we observe that finetuning on our proposed dataset often yields more significant improvement compared to finetuning on the corresponding benchmark's data. In addition, it results in only a slight degradation on popular benchmarks such as MMLU \citep{hendrycks2021measuring} and HellaSwag \citep{zellers2019hellaswag}, indicating that the overall capabilities of the models remain largely unaffected. Finally, another advantage of our proposed dataset is that it contains no \emph{factual} information; as it was recently discovered by \citet{gekhman2024does}, finetuning on previously unseen knowledge may encourage hallucinations. Thus, finetuning on our key-value dataset improves LLMs' retrieval and reasoning without suffering from such unwanted characteristics.

\par Our findings highlight the potential of finetuning on synthetic data as a promising approach to enhancing the performance of LLMs on real downstream tasks. Our paper is organized as follows: in Section \ref{sec:synthetic_data} we describe the format of the proposed dataset, and its variations that provide (or not) an answer template to the model, in Section \ref{sec:exp_res} we present our experimental results, in Section \ref{sec:limitations} we discuss the main limitations and possible future directions of our work, and in Section \ref{sec:conclusion} we discuss our main conclusions.

\begin{figure}[t]
\centering
\begin{tikzpicture}
    \node[draw, rounded corners, text width=1\linewidth, align=left, inner xsep=.5em, inner ysep=1.3em] (box) {  \small\tt
        Do a task using the list of dictionaries below.\\
        \hfill\\
        Dictionary [1] \{122: 765, 4548: 1475, 4818: 4782\}\\
        Dictionary [2] \{526: 290, 9205: 9318, 9278: 1565\}\\
        ...\\
        Dictionary [32] \{2931: 8364, 196: 1464, 812: 5363\}\\
        ...\\
        Dictionary [85] \{344: 1579, 116: 617, 330: 411\}\\
        \hfill\\
        Above is a list of dictionaries such that each key and value is an integer. Report the value of key 2931 and the dictionary it is in.\\
        \hrulefill\\
        Desired answer: The value of key 2931 is 8364 and it is in Dictionary [32].
    };
    \node[fill=black, text=white, rounded corners, yshift=0em] at (box.north) {Simple dictionary key-value retrieval};
  \end{tikzpicture}
\caption{An example prompt with desired answer of simple dictionary key-value retrieval task.}
\label{fig:simpledict_ntemplate}
\end{figure}

\subsection{Related work}
\paragraph{Long Context LLMs.} Recent works have observed LLMs' limited retrieval and reasoning capabilities in the long-context setting. \citet{liu2023lost} discovered a positional bias when LLMs retrieve information from long contexts. In particular, the authors found out that the retrieval accuracy drops when the desired information lies in the middle of the context. \citet{gkamradt2023needle} conducted the ``needle-in-a-haystack'' experiment by placing a random fact (the ``needle'') in a long input context (the ``haystack'') and observed that LLMs struggle to spot the needle as the input context length grows. To mitigate this behavior, \citet{yu2024training} and \citet{an2024make} finetuned LLMs on long-context augmentation data consisting of long-context question-answering tasks to enhance LLMs' long-context capabilities. \citet{tang2023found} shuffled the prompt and marginalized the prompt order biases in the long-context setting and \citet{zhang2024found} re-scaled the indices in positional encoding. \citet{levy2024same} introduced a benchmark, FLenQA, by extending input samples with varying lengths and types of padding, discovering LLMs' significant degradation in reasoning ability at context lengths much shorter than the maximum limit. \par
There are also other relevant works on long-context LLMs \citep{junqing2023never,mohtashami2023landmark,chen2023extending,bai2023longbench,an2023eval}. \citet{xu2023retrieval} showed that Retrieval Augmented Generation (RAG) can be as accurate as full finetuning on longer context windows. \citet{chen2023walking} extended the LLM's predetermined context limit by treating it as an interactive agent who processes the input through iterative prompting. \citet{jinllm} extended LLM's context window by remapping the unseen relative positions during inference. \citet{fu2024data} proposed a data engineering recipe for scaling LLMs to $128k$ context lengths through lightweight continual pretraining on a balanced mixture of length-upsampled data. \citet{peysakhovich2023attention} proposed ``attention sorting,'' a method that improves long context models by iteratively sorting documents based on attention and generating responses with the re-ordered context.

\paragraph{Data-centric AI.} In recent years, the field of data-centric AI has emerged, which focuses on improving the quality and efficiency of AI systems through data-oriented approaches rather than model-centric techniques \citep{sener2018active, ghorbani2019data, zha2023data, albalak2024survey}. \citet{gadre2024datacomp} and \citet{mazumder2024dataperf} proposed benchmarks that fix model training code, where the goal is to design better datasets to achieve better performance. \citet{lee2023teaching} and \citet{zhou2024transformers} studied the data format in training transformers to learn arithmetic tasks.

\paragraph{LLM Benchmarks and Evals.}
Much research has been recently conducted towards the design of meaningful benchmarks that probe the capabilities of LLMs. Benchmarks such as GLUE \citep{wang2018glue}, SuperGLUE \citep{wang2019superglue} test whether a model has general language understanding capabilities. MMLU \citep{hendrycks2021measuring} aims to measure the models' accuracy across a wide variety of tasks that span STEM, humanities, social sciences, and more, while GSM8k \citep{cobbe2021training} tests capabilities on school math. In HellaSwag \citep{zellers2019hellaswag} models are presented with an event description and must select the most likely follow-up sentence from a set of carefully selected choices, while HumanEval \citep{chen2021evaluating} measures their ability to generate code given docstrings. TriviaQA \citep{joshi2017triviaqa} is a reading comprehension benchmark and NQ-Open \citep{lee-etal-2019-latent, tacl_a_00276} is an open domain question-answering benchmark where the question-answer pairs are collected from a diverse set of fields.

\section{Synthetic dataset of retrieval tasks}
\label{sec:synthetic_data}
In this section, we introduce the dataset on which we finetune the models. The dataset consists of two synthetic retrieval tasks: 1) simple dictionary key-value retrieval and 2) multi-subkey dictionary key-value retrieval. 

\paragraph{Simple dictionary key-value retrieval.}
In this task, we provide the model with a list of dictionaries of integer keys and values, and ask it to retrieve the value of a specified key (denoted as the \emph{gold key}). Figure~\ref{fig:simpledict_ntemplate} shows an example  of this task.

\begin{figure}[t]
\centering
\begin{tikzpicture}
    \node[draw, rounded corners, text width=1\linewidth, align=left, inner xsep=.5em, inner ysep=1.3em] (box) {  \small\tt
    Do a task using the list of dictionaries below.\\
    \hfill\\
    Dictionary [1] \{(141, 986, 163): 2528, (726, 947, 349, 820): 4130\}\\
    Dictionary [2] \{(555, 710, 424): 5756, (623, 141, 997): 1633, (957, 634, 969): 7871\}\\
    ...\\
    Dictionary [6] \{(645, 417, 847): 6409, (141, 623, 616): 5617\}\\
    ...\\
    Dictionary [49] \{(710, 105, 141, 799): 5369, (623, 210, 477): 8971, (899, 126, 999): 4409\}\\
    \hfill\\
    Above is a list of dictionaries such that each key is a tuple of integers and each value is an integer. Report the key that contains the integers 616, 141, 623 (not necessarily in order), its value, and the dictionary it is in.\\
    \hrulefill\\
    Desired answer: The key that contains the integers 616, 141, 623 is (141, 623, 616). Its value is 5617 and it is in Dictionary [6].
    };
    \node[fill=black, text=white, rounded corners, yshift=0em] at (box.north) {Multi-subkey dictionary key-value retrieval};
  \end{tikzpicture}
\caption{An example prompt with desired answer of multi-subkey dictionary key-value retrieval task. Here \texttt{(141, 623, 616)} is the \emph{gold key}. Note that \texttt{141} and \texttt{623} in the \emph{gold key} are also subkeys of other keys.}
\label{fig:multsubkeydict_ntemplate}
\end{figure}

\begin{figure}[t]
\centering
\begin{tikzpicture}
    \node[draw, rounded corners, text width=1\linewidth, align=left, inner xsep=.5em, inner ysep=1.3em] (box) { \small \tt
        Do a task using the list of dictionaries below.\\
        \hfill\\
        Dictionary [1] \{122: 765, 4548: 1475, 4818: 4782\}\\
        Dictionary [2] \{526: 290, 9205: 9318, 9278: 1565\}\\
        ...\\
        Dictionary [32] \{2931: 8364, 196: 1464, 812: 5363\}\\
        ...\\
        Dictionary [85] \{344: 1579, 116: 617, 330: 411\}\\
        \hfill\\
        Above is a list of dictionaries such that each key and value is an integer. Report the value of key 2931 and the dictionary it is in. Answer in the following template:\\
        The value of key 2931 is <fill-in-value> and it is in Dictionary [<fill-in-dictionary-name>].\\
        \hrulefill\\
        Desired answer: The value of key 2931 is 8364 and it is in Dictionary [32].
    };
    \node[fill=black, text=white, rounded corners, yshift=0em] at (box.north) {Simple dictionary key-value retrieval (with an answer template)};
  \end{tikzpicture}
\caption{The prompt of the simple dictionary key-value retrieval task is provided with an answer template.}
\label{fig:simpledict_wtemplate}
\end{figure}

\begin{figure}[t]
    \centering
    \includegraphics[width=1\textwidth]{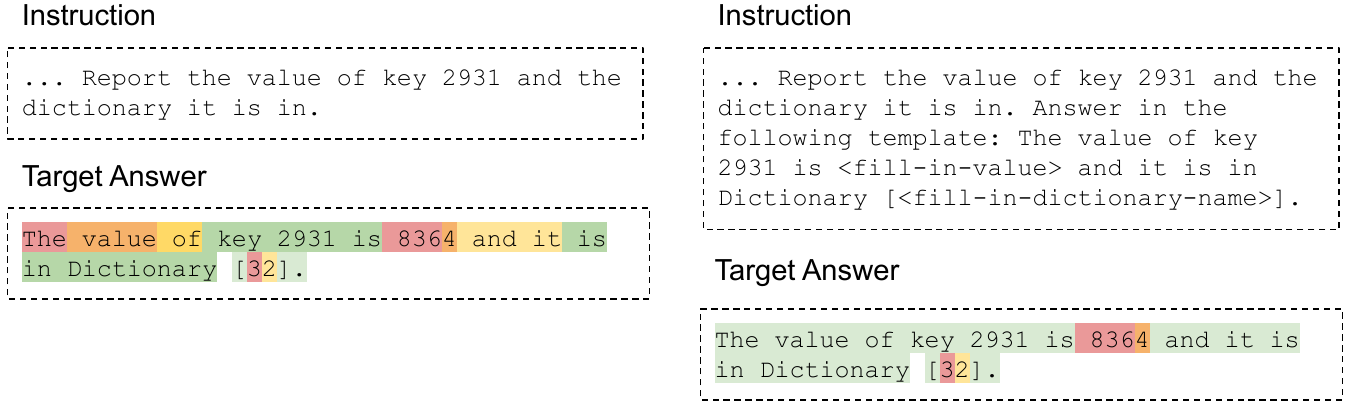}
    \caption{Token-level loss on the target answer when provided with (right) and without (left) an answer template, where red indicates high and green low loss.}
    \label{fig:template_loss}
\end{figure}

\paragraph{Multi-subkey dictionary key-value retrieval.}
For models that can already tackle the first task (e.g., for the first task GPT 3.5 Turbo achieves around $0.99$ accuracy irrespective of the position of \emph{gold key}), we design a harder version of the key-value retrieval task where each key is a tuple of subkeys. Other keys can share some but not all of the subkeys of the \emph{gold key}. We increase the difficulty of this task by randomizing the order of subkeys in the prompt so that the order is not necessarily the same as that of the \emph{gold key}. Figure~\ref{fig:multsubkeydict_ntemplate} shows an example of this task.
\paragraph{Prompt with an answer template.} 
Note that with the prompt in Figure 
\ref{fig:simpledict_ntemplate}, slightly different answers like ``\texttt{8364} is the value of key \texttt{2931} in dictionary \texttt{32}'' and ``Dictionary [\texttt{32}] has the key \texttt{2931} with value \texttt{8364}'' are also correct. Therefore, since the model is finetuned on the entire answer, during supervised finetuning, it also learns the format of our provided answer besides learning to retrieve the desired value. In order to make the model only focus on retrieving the correct value without being affected by the format of the answer, we provide the model with an answer template with which we want the model to answer. Figure \ref{fig:simpledict_wtemplate} shows an example of a prompt with an answer template. In Figure \ref{fig:template_loss} we visualize the token-level loss on the target answer, where red indicates high and green low loss. If an answer template is provided, the loss on the formatting part is small. This lets the model to focus on the important part and learn the right skill rather than how to answer the question.

\section{Experiments and results}

\label{sec:exp_res}
Our goal is to investigate whether finetuning LLMs (in particular, GPT-3.5 Turbo and Mistral 7B \footnote{\texttt{gpt-3.5-turbo-1106} and \texttt{Mistral-7B-Instruct-v0.1}}) on our proposed synthetic numerical retrieval tasks improves their long context capabilities on natural language tasks: multi-document question answering (MDQA) \citep{liu2023lost} and flexible length question answering (FLenQA) \citep{levy2024same}.

\subsection{Stage 1: Finetuning LLMs on synthetic retrieval tasks}
For Mistral 7B, our dataset consists of $350$ samples of simple dictionary key-value retrieval tasks. Each task has $85$ dictionaries and each dictionary has $3$ to $4$ keys, so each prompt has roughly $3900$ tokens (to leave space for the tokens in the answer as \texttt{Mistral-7B-Instruct-v0.1} uses a sliding window context length of $4096$). We finetune the model on only the answer part (masking out the instruction part) for $2$ epochs. More implementation details are in \ref{sec:mistral_ft}. Figure \ref{fig:mistral-simpledict} shows Mistral 7B's performance on simple dictionary key-value retrieval task before and after finetuning.\par
Since GPT-3.5 Turbo already performs well on simple dictionary key-value retrieval task, we finetune it on multi-subkey dictionary key-value retrieval tasks. The dataset consists of $150$ samples and each sample has $49$ dictionaries. We finetune the model for $3$ epochs using OpenAI's API.

\subsection{Stage 2: Evaluations on long context retrieval and reasoning tasks}
\subsubsection{Multi-document question answering (MDQA)}
\begin{figure}[t]
    \centering
    \begin{subfigure}[ht]{0.49\textwidth}
        \centering
        \includegraphics[width=1\textwidth]{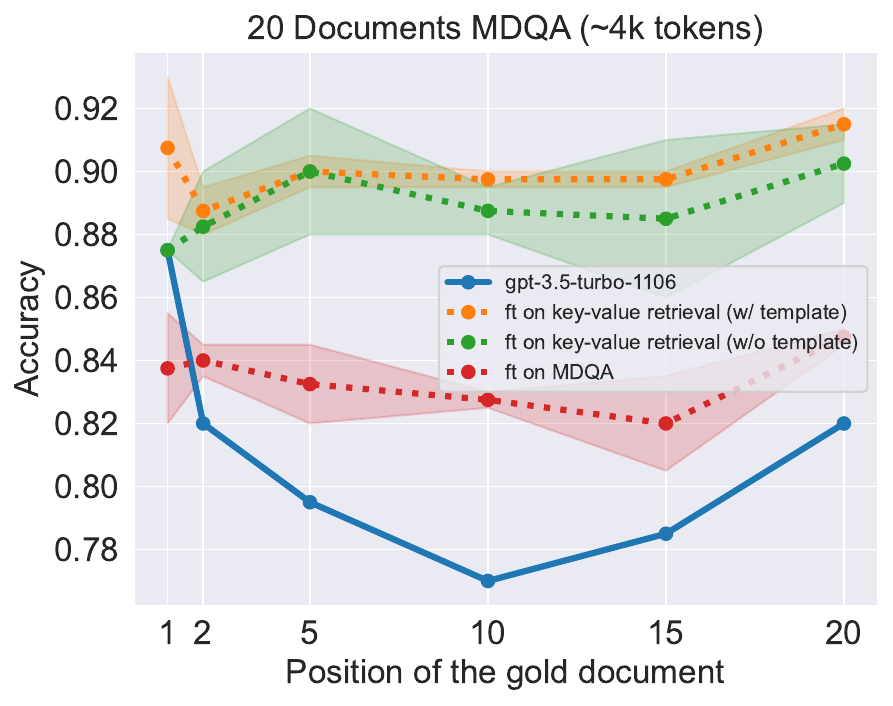}
        \caption{GPT-3.5 Turbo and the finetuned versions.}
        \label{fig:gpt-mdqa}
    \end{subfigure}
    ~
    \begin{subfigure}[ht]{0.49\textwidth}
        \centering
        \includegraphics[width=1\textwidth]{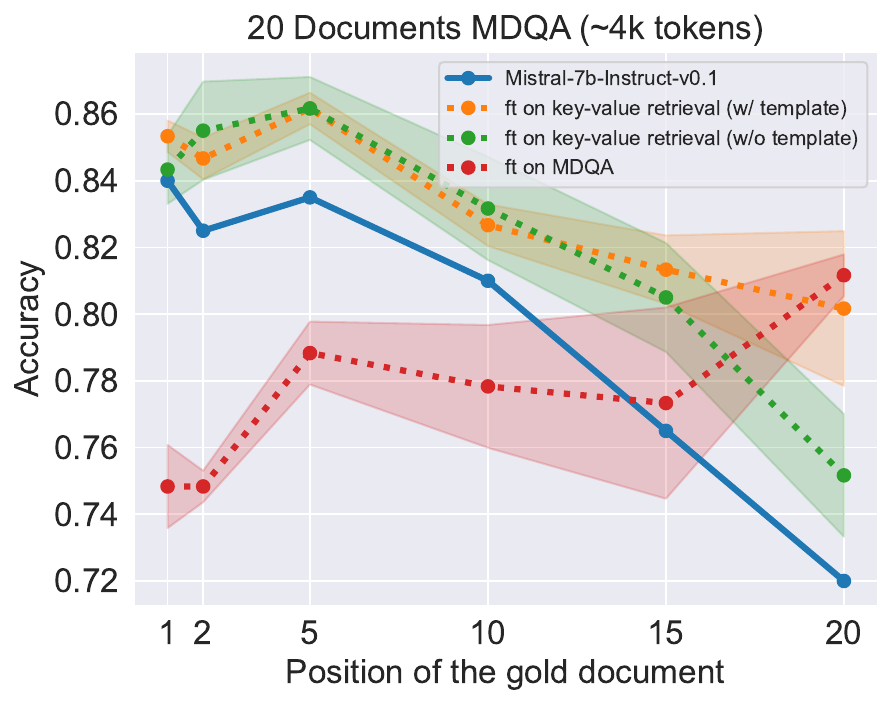}
        \caption{Mistral 7B and the finetuned versions.}
        \label{fig:mistral-mdqa}
    \end{subfigure}
    \caption{Performance of GPT-3.5 Turbo, Mistral 7B and their corresponding finetuned versions on the MDQA task.}
    \label{fig:mdqa}
\end{figure}

We test models' capabilities of retrieving important information in a long context setting. In MDQA, we provide the model with $k$ documents and prompt it to answer a question such that only $1$ of $k$ documents (denoted as the \emph{gold document}) contains the answer and the other $k-1$ documents (denoted as \emph{distractors}) are completely irrelevant to the question. We test the setting of a context with $20$ documents (around 4K tokens) and place \emph{gold document} at positions $\{1, 2, 5, 10, 15, 20\}$ \footnote{For example, \emph{gold document} placed at position $1$ means it is the first document in the context.}. For each position, we test the model on $200$ task samples and measure the accuracy using the maximum subspan exact match as in \citep{liu2023lost}.\par
\begin{highlight}
    \paragraph{Finding 1:} 
    \emph{Finetuning LLMs on synthetic key-value retrieval tasks enhances their performance on practical retrieval tasks, demonstrating effective transfer of learned capabilities.}
\end{highlight}
The result of $20$ documents MDQA is shown in Figure~\ref{fig:mdqa}, where x-axis is the position of \emph{gold document}. In Figure~\ref{fig:gpt-mdqa}, for the original GPT-3.5 Turbo model, there is a U-shaped performance curve, indicating that the performance is highest if the important information is at the beginning or at the end of the input context, with the model struggling to retrieve the answer if the important information is in the middle. Finetuning the models on synthetic retrieval tasks flattens the U-shaped curve and information is much more accurately retrieved over all positions across the input context. In Figure~\ref{fig:mistral-mdqa}, the original Mistral 7B model has a primacy bias -- in the sense that it can more accurately retrieve information that is at the beginning of the input context. Finetuning the models on our proposed data mitigates this primacy bias, and hence, information at the end of the context is much more accurately retrieved.\par
\begin{highlight}
    \paragraph{Finding 2:} 
    \emph{Synthetic data is better than MDQA data even if the goal is to perform better in MDQA task.}
\end{highlight}
As a comparison, we also finetune the models on the MDQA dataset itself for roughly the same number of training tokens and see how finetuned models perform. Since the MDQA dataset only provides the ground truth answers in one or two words, we prompt GPT-3.5 Turbo with correct answers and let it form a complete sentence as the target answer. As shown in Figure \ref{fig:gpt-mdqa}, GPT-3.5 Turbo finetuned on our synthetic data perform better than the one finetuned on MDQA. In Figure \ref{fig:mistral-mdqa} we can see that despite training on MDQA tasks, Mistral 7B still struggles to perform well on MDQA, with a significant performance drop when \emph{gold document} is at the beginning of the prompt. These findings underscore the effectiveness of our synthetic data generation method, which  enhances performance on specific datasets like MDQA, even surpassing direct finetuning on the target dataset.

\subsubsection{Flexible length question answering (FLenQA)}
\label{sec:flenqa-res}
\begin{figure}[ht]
    \centering
    \begin{subfigure}[ht]{0.49\textwidth}
        \centering
        \includegraphics[width=1\textwidth]{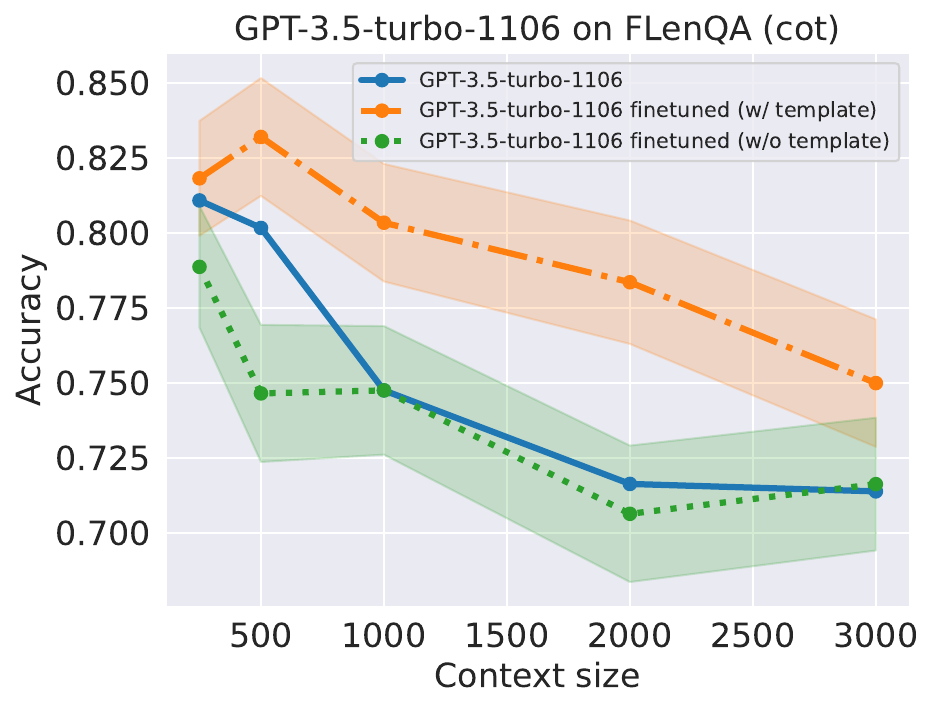}
        \caption{GPT-3.5 Turbo and the finetuned versions.}
        \label{fig:gpt-flenqa-cot}
    \end{subfigure}
    ~ 
    \begin{subfigure}[ht]{0.49\textwidth}
        \centering
        \includegraphics[width=1\textwidth]{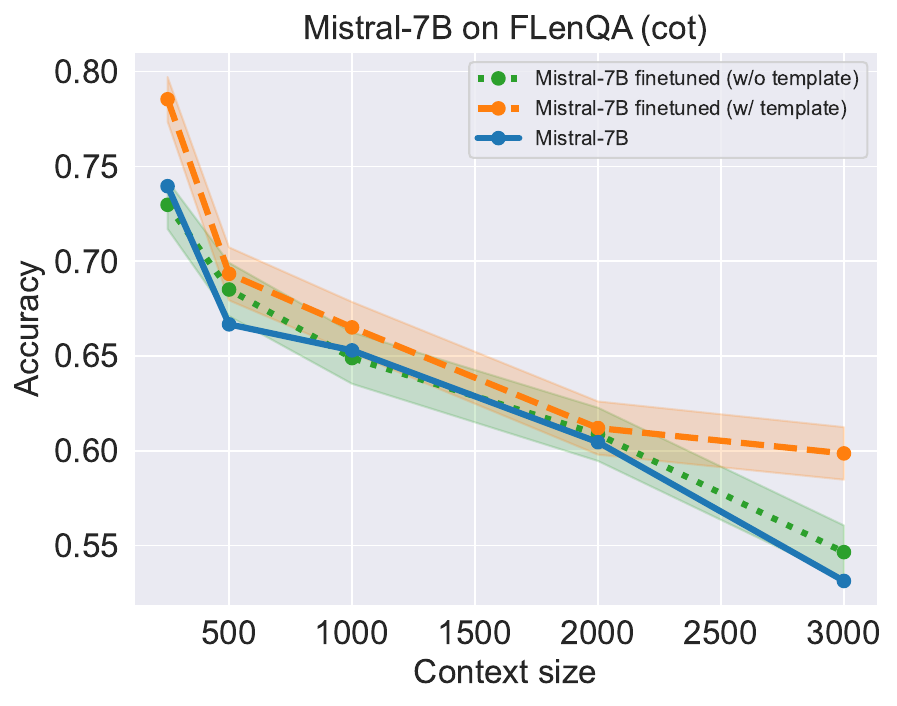}
        \caption{Mistral 7B and the finetuned versions.}
        \label{fig:mistral-flenqa-cot}
    \end{subfigure}
    \caption{Performance of GPT-3.5 Turbo, Mistral 7B and their corresponding finetuned versions on the FLenQA task, using chain-of-thought prompting.}
    \label{fig:flenqa-cot}
\end{figure}
\begin{figure}[ht]
    \centering
    \begin{subfigure}[ht]{0.49\textwidth}
        \centering
        \includegraphics[width=1\textwidth]{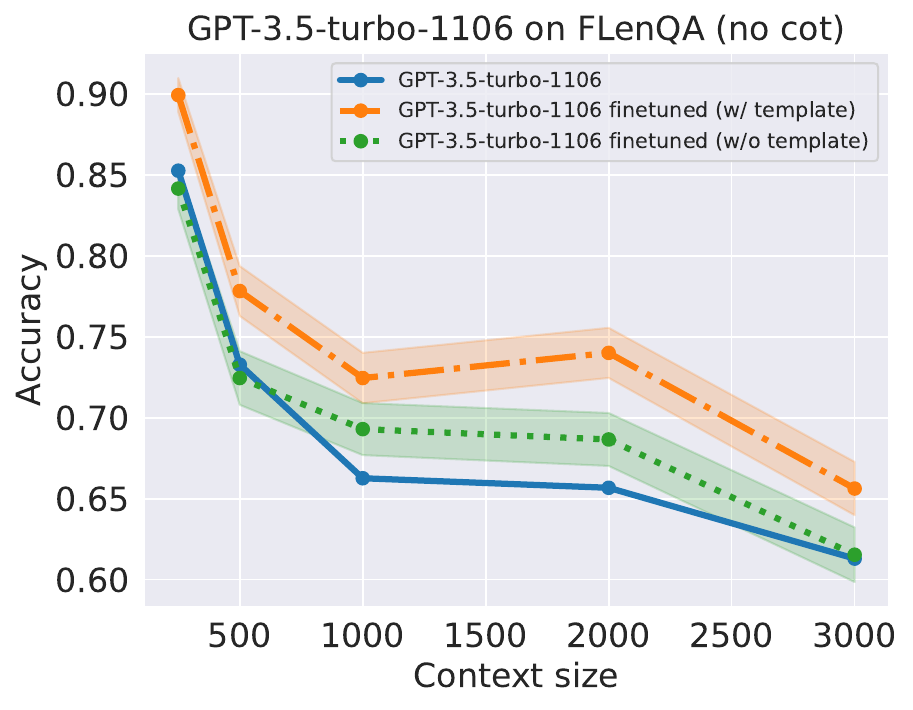}
        \caption{GPT-3.5 Turbo and the finetuned versions.}
        \label{fig:gpt-flenqa}
    \end{subfigure}
    ~ 
    \begin{subfigure}[ht]{0.49\textwidth}
        \centering
        \includegraphics[width=1\textwidth]{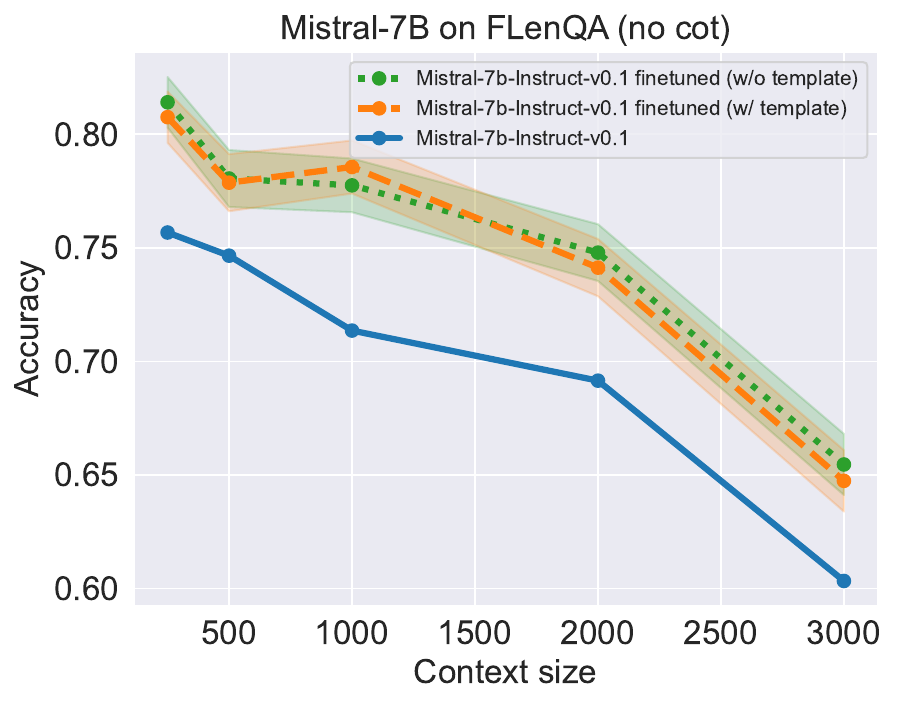}
        \caption{Mistral 7B and the finetuned versions.}
        \label{fig:mistral-flenqa}
    \end{subfigure}
    \caption{Performance of GPT-3.5 Turbo, Mistral 7B and their corresponding finetuned models on the FLenQA task without employing chain-of-thought prompting. }
    \label{fig:flenqa}
\end{figure}

We also test models' long context reasoning capabilities. FLenQA is a dataset comprising reasoning tasks with varying length that ranges from $250$ tokens to $3000$ tokens. Each task consists of a context and a ``True'' or ``False'' question that can be answered by two key sentences from the context. We test chain-of-thought \citep{wei2022chain} and non chain-of-thought prompting, each with a total of $2000$ task samples. For chain-of-thought prompting, we ask the model to produce the result step by step and derive the answer (``True'' or ``False'') at the end, and in the non chain-of-thought prompting we ask the model to directly answer ``True'' or ``False''.\par
\begin{highlight}
    \paragraph{Finding 3:} 
    \emph{Finetuning LLMs on synthetic key-value retrieval tasks improves LLMs' long-context reasoning capabilities, even if explicit chain-of-thought reasoning is not allowed.}
\end{highlight}
In Figure \ref{fig:flenqa-cot} and \ref{fig:flenqa} we present our results on the FLenQA dataset. The x-axes represent the number of tokens in the context, while the y-axes represent the accuracy of the response. Figure \ref{fig:flenqa-cot} shows results where chain-of-thought prompting is employed. In Figure~\ref{fig:gpt-flenqa-cot}, we notice that although the model suffers from a performance drop if finetuned on data without answer template, finetuning GPT-3.5 Turbo on data with answer template significantly improves model's chain-of-thought reasoning capability. In Figure~\ref{fig:mistral-flenqa-cot} we can also see that finetuning Mistral 7B on data with answer template improves models chain-of-thought capability. We hypothesize that the reason for this is that the finetuned models utilize their improved retrieval capabilities to capture relevant information more accurately, which helps them deduce the answer.

\par Figure \ref{fig:flenqa} presents results where models are required to directly answer with ``True" or ``False" without providing explicit reasoning. The results show a notable improvement in performance for finetuned models. This improvement is significant because it demonstrates that, even if explicit reasoning (that is related to retrieval capability) is not allowed, finetuning on our proposed synthetic tasks enhances the models' internal reasoning capabilities.

\par
\begin{highlight}
    \paragraph{Finding 4:} 
    \emph{LLMs finetuned on synthetic tasks with answer templates are better.}
\end{highlight}
From Figure \ref{fig:mdqa}, \ref{fig:flenqa-cot} and \ref{fig:flenqa}, we can observe that models finetuned on synthetic key-value retrieval tasks with answer templates perform better on MDQA and FLenQA than that on without answer templates. This verifies our hypothesis that having an answer template helps the model learn the right skill more efficiently. This highlights a key advantage of synthetic data: it allows for greater control over the model's output format. Unlike real-world tasks where developing answer templates can be challenging, synthetic tasks allow for easy implementation of structured response formats, facilitating skill learning.

%============================

\subsection{Stage 3: Evaluation of finetuned models' general capabilities}
\begin{highlight}
    \paragraph{Finding 5:} 
    \emph{Finetuning LLMs on synthetic key-value retrieval tasks does not hurt models' general capabilities.}
\end{highlight}
One possible drawback of our approach is that finetuning on the proposed artificial tasks would severely harm the general purpose capabilities of the tested models. In order to assess this concern, we tested the original and finetuned versions of GPT-3.5 Turbo and Mistral 7B on some general purpose benchmarks. Note that for our assessments we used the codebases of \citet{eval-harness} and \citet{fu2023chain}.

\begin{table}[ht]
\centering
\resizebox{\textwidth}{!}{
\begin{tabular}{cccccc}
\toprule
\textbf{MODEL} &\textbf{MMLU} &\textbf{HellaSwag} &\textbf{GSM8K} &\textbf{Triviaqa} & \textbf{NQ-Open}\\
\midrule
Mistral-7B &53.42 &56.31 &34.65 & 47.63 & 11.61\\
Mistral-7B ft (w/template) &53.44 ($+0.02$) &56.22 ($-0.09$) &34.34 ($-0.31$) &47.74 ($+0.11$) & 11.98 ($+0.37$)\\
Mistral-7B ft (w/o template)&53.42 ($-0.00$) &56.30 ($-0.01$) & 34.14 ($-0.51$) &47.62 ($-0.01$) & 11.40 ($-0.21$)\\
\midrule
GPT-3.5-turbo & 68.07 & - & 72.33 & - & -\\
GPT-3.5-turbo ft (w/template) & 67.75 ($-0.32$)& - & 71.65 ($-0.68$) & - & -\\
GPT-3.5-turbo ft (w/o template) & 68.16 ($+ 0.09$) & - & 75.06 ($+ 2.73$) & - & -\\
\bottomrule
\end{tabular}}
\caption{Model's performance evaluated on general ability benchmarks. All numbers are reported in percentage. Here ``w/'' and ``w/o'' denote the models that are finetuned on the the synthetic tasks that were described in Section \ref{sec:synthetic_data}.}
\label{table:general-benchmarks}
\end{table}

The results can be seen in Table~\ref{table:general-benchmarks}. In particular, we consider five widely used benchmarks: MMLU \citep{hendrycks2021measuring}\footnote{Due to computational constraints, we did not evaluate  GPT-3.5 Turbo on all benchmarks, and for MMLU we use 20\% of the full dataset.}, HellaSwag \citep{zellers2019hellaswag}, GSM8k \citep{cobbe2021training}, TriviaQA \citep{joshi2017triviaqa} and NQ-Open \citep{47761}. What we can observe is that all the finetuning strategies result in no significant degradation on the general purpose benchmarks mentioned above.

\subsection{Stage 4: Comparisons with other baselines}
We also consider three additional long-context augmentation datasets as baselines: MultidocQA \citep{yu2024training}, IN2 \citep{an2024make}, and Needle-in-a-haystack \citep{gkamradt2023needle}. MultidocQA is a dataset of multiple documents question and answering where the model needs to paraphrase the document before answering. IN2 is a long-context question answering dataset where the answer can be deduced from one or multiple parts of the context. Needle-in-a-haystack is a widely used long-context test set where the model is prompted to identify some key information (the needle) within a long context (the haystack). We finetune Mistral 7B on these baselines, using roughly the same number of training tokens and report their performance on MDQA, FLenQA, and general purpose benchmarks.

\begin{figure}[ht]
    \centering
    \begin{subfigure}[ht]{0.49\textwidth}
        \centering
        \includegraphics[width=1\textwidth]{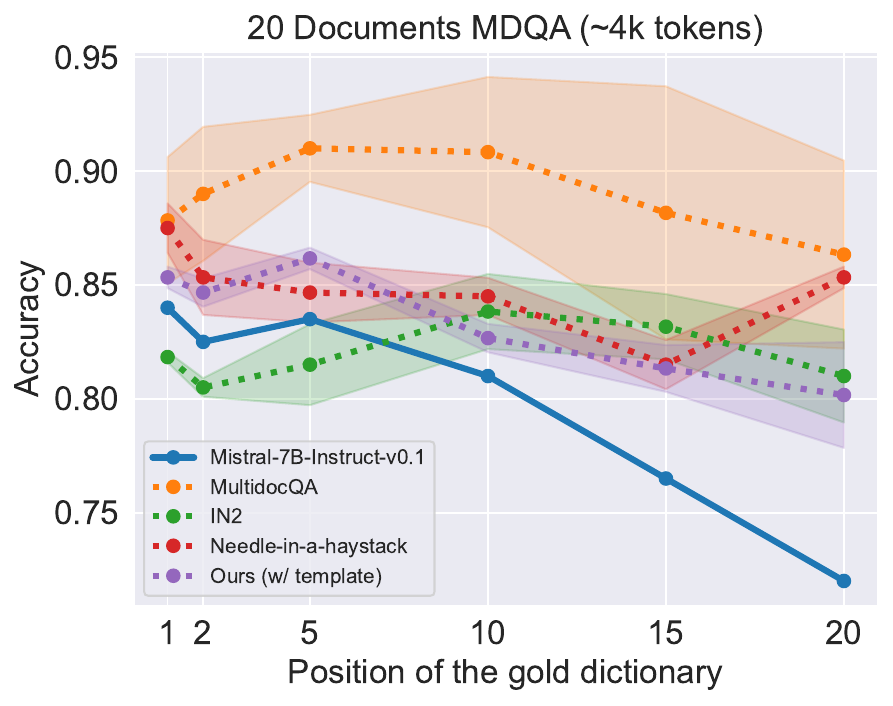}
        \caption{MDQA}
        \label{fig:mdqa-baselines}
    \end{subfigure}
    ~ 
    \begin{subfigure}[ht]{0.49\textwidth}
        \centering
        \includegraphics[width=1\textwidth]{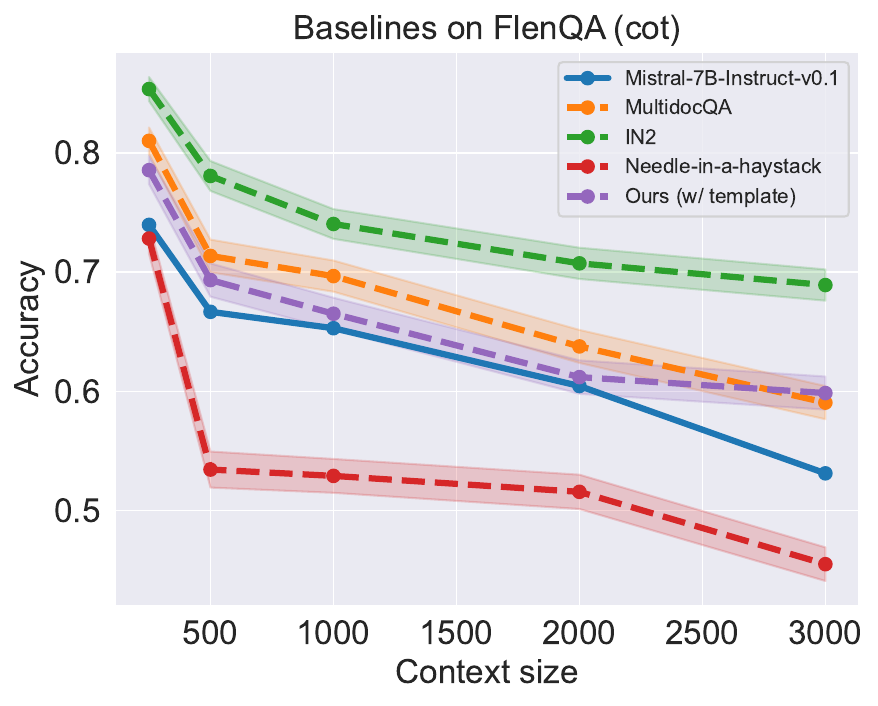}
        \caption{FLenQA with chain-of-thought prompting}
        \label{fig:flenqa-cot-baselines}
    \end{subfigure}
    ~
    \begin{subfigure}[ht]{0.49\textwidth}
        \centering
        \includegraphics[width=1\textwidth]{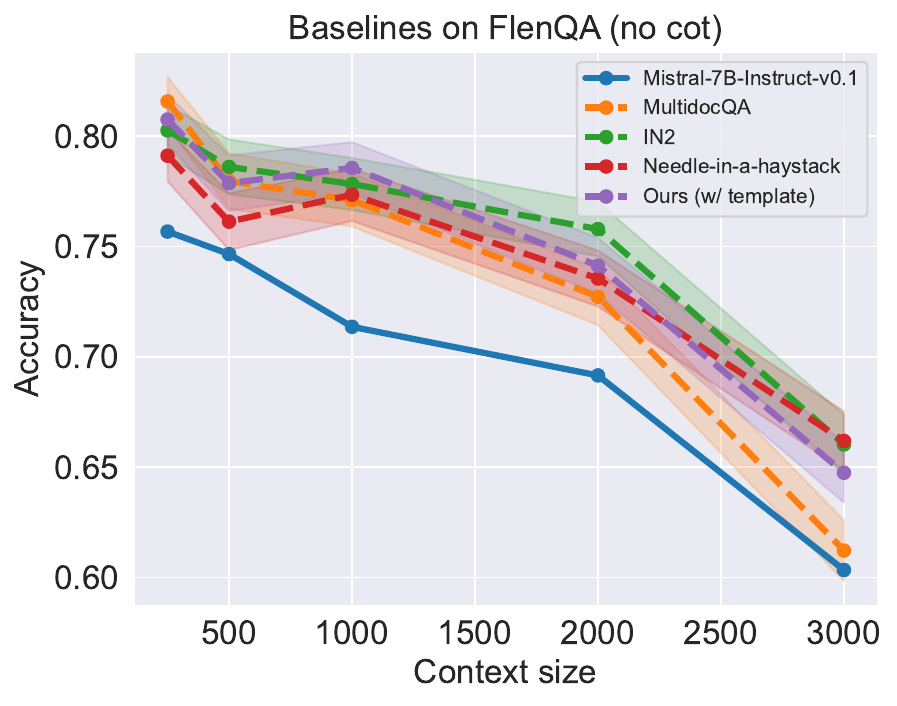}
        \caption{FLenQA without chain-of-though prompting}
        \label{fig:flenqa-nocot-baselines}
    \end{subfigure}
    \caption{Performance of finetuned Mistral 7B on (a) MDQA, (b) FLenQA with chain-of-thought prompting, and (c) FLenQA without chain-of-thought prompting.}
    \label{fig:baselines}
\end{figure}

\begin{table}[ht]
\centering
\scriptsize
\resizebox{\textwidth}{!}{
\begin{tabular}{cccccc}
\toprule
\textbf{Finetuning dataset} &\textbf{MMLU} &\textbf{HellaSwag} &\textbf{GSM8K} &\textbf{Triviaqa} & \textbf{NQ-Open}\\
\midrule
Original Mistral-7B &53.42 &56.31 &34.65 & 47.63 & 11.61\\
Ours (w/template) &53.44 ($+0.02$) &56.22 ($-0.09$) &34.34 ($-0.31$) &47.74 ($+0.11$) & 11.98 ($+0.37$)\\
MultidocQA \citep{yu2024training} & 53.19 (-0.22) & 56.27 (-0.04) & 33.28 (-1.36) & 45.20 (-2.43) & 8.69 (-2.91) \\
IN2 \citep{an2024make} & 53.49 (+0.07) & 56.44 (+0.13) & 34.98 (+0.32) & 45.44 (-2.19) & 9.80 (-1.81) \\
Needle-in-a-haystack \citep{gkamradt2023needle} & 52.83 (-0.59) & 56.22 (-0.09) & 33.79 (-0.86) & 41.30 (-6.33) & 4.88 (-6.73) \\
\bottomrule
\end{tabular}}
\caption{Mistral 7B and finetuned versions' performance evaluated on general ability benchmarks. All numbers are reported in percentage.}
\label{table:general-benchmarks-baselines}
\end{table}

\begin{highlight}
    \paragraph{Finding 6:} 
    \emph{Synthetic data do not encourage hallucinations that other baselines may yield.}
\end{highlight}
From Figure \ref{fig:baselines} and Table \ref{table:general-benchmarks-baselines}, we can see that while some baselines outperform our proposed data on either MDQA or FLenQA, they all have more significant degradation on the general benchmarks we test, especially on TriviaQA and NQ-Open. One possible reason is that all other baselines contain factual information. \citet{gekhman2024does} shows that finetuning on factual information encourages hallucinations, something that we verify observing the significant degradation on TriviaQA and NQ-Open, which are knowledge-based benchmarks. In contrast, our proposed dataset is purely synthetic, comprising of key-value pairs, and as a result, does not encourage hallucinations. We also highlight another benefit of our synthetic data: since it does not contain any factual information, it will not have the problem of containing potential outdated information that further encourages hallucinations, from which other long-context augmentation datasets may suffer.

\subsection{Stage 5: Evaluation on longer-context setting}
\begin{figure}[h]
    \centering
    \includegraphics[width=0.5\textwidth]{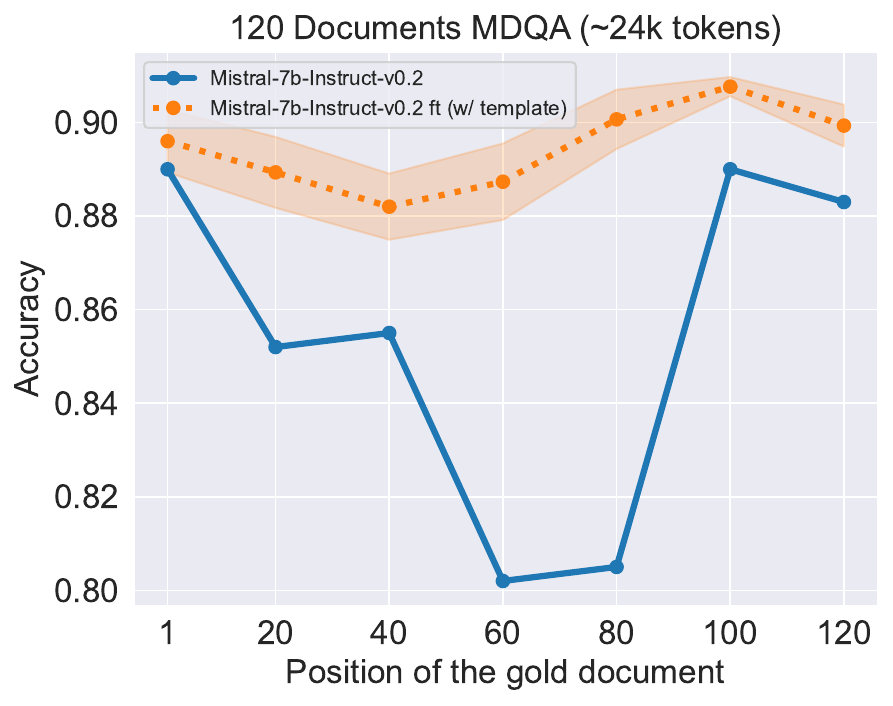}
    \caption{Performance of finetuned \texttt{Mistral-7b-Instruct-v0.2} on 120 documents MDQA.}
    \label{fig:longer-context}
\end{figure}

We also test the longer-context setting. We finetune \texttt{Mistral-7b-Instruct-v0.2} on simple key-value retrieval task with maximum context length of 24K and test it on MDQA. We observe a clear improvement over the original model as shown in Figure \ref{fig:longer-context}.

\section{Limitations and future work} \label{sec:limitations}
\begin{figure}[t]
    \centering
    \includegraphics[width=0.6\textwidth]{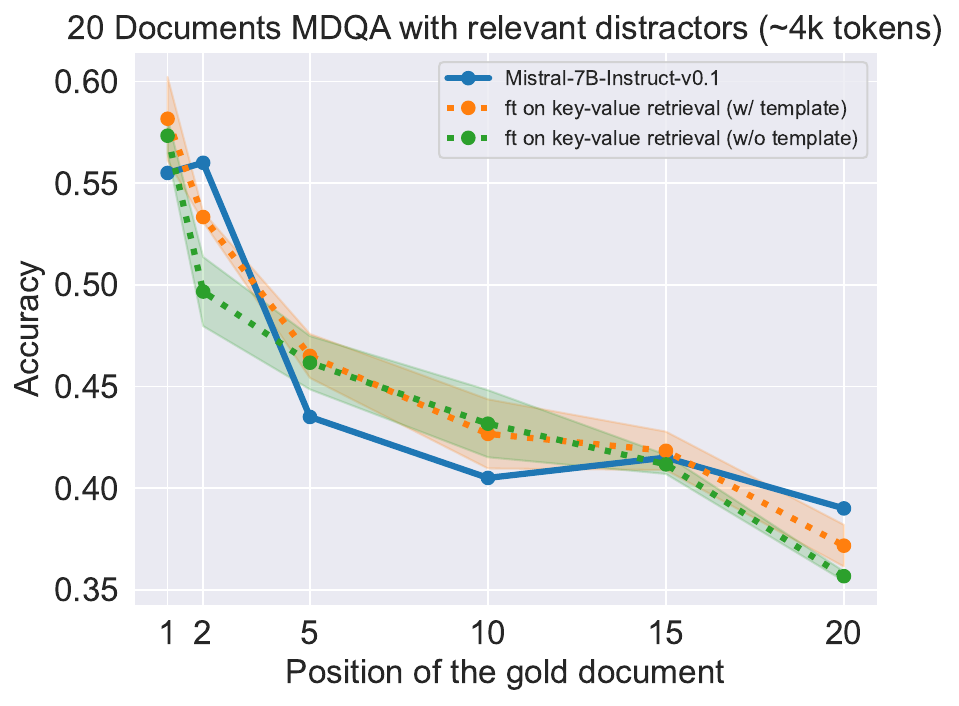}
    \caption{Mistral 7B and the finetuned versions on MDQA with relevant distractors. The finetuned variants do not show a significant improvement over the original model.}
    \label{fig:mistral-mdqa-relevant}
\end{figure}

Our dataset does have a limitation. MDQA benchmark also has another version where \emph{distractors} are relevant distractors, meaning that they are documents retrieved by a retrieval system (based on the relevance score) that do not contain the answer. Models finetuned on our dataset will not improve in this setting, as is shown in Figure \ref{fig:mistral-mdqa-relevant}. A possible future work of this study is to add our synthetic retrieval dataset as a small part of a larger instruction finetuning dataset and see the difference between models finetuned with and without synthetic retrieval data and observe how they perform differently on long context retrieval and reasoning tasks.

\section{Conclusion} \label{sec:conclusion}
In this work, we introduce a novel finetuning approach that leverages carefully designed synthetic datasets to enhance the information retrieval and reasoning capabilities of LLMs in real downstream tasks. Our study demonstrates that finetuning on our proposed synthetic data significantly improves the performance of the tested models on tasks like MDQA and FLenQA, mitigating the ``lost-in-the-middle" behavior that was observed in \citet{liu2023lost}. On the other hand, we find that after finetuning, the models' performance on general benchmarks remains almost constant, something that indicates that their overall capabilities are mostly unaffected. We also find that compared to other long-context augmentation datasets that contain factual information, our purely artificial data does not encourage hallucinations. Moreover, it will not have the problem of containing potential outdated information. Thus, we believe that our study demonstrates the potential of finetuning LLMs on carefully crafted synthetic datasets to enhance their capabilities on downstream tasks. We hope that our findings will inspire further research into the development of effective synthetic datasets.

\bibliography{arxiv}
\bibliographystyle{abbrvnat}
\newpage

\appendix
\section{Appendix}
\subsection{Finetuning Mistral 7B and GPT 3.5 Turbo}
\begin{figure}[h]
    \centering
    \includegraphics[width=0.6\textwidth]{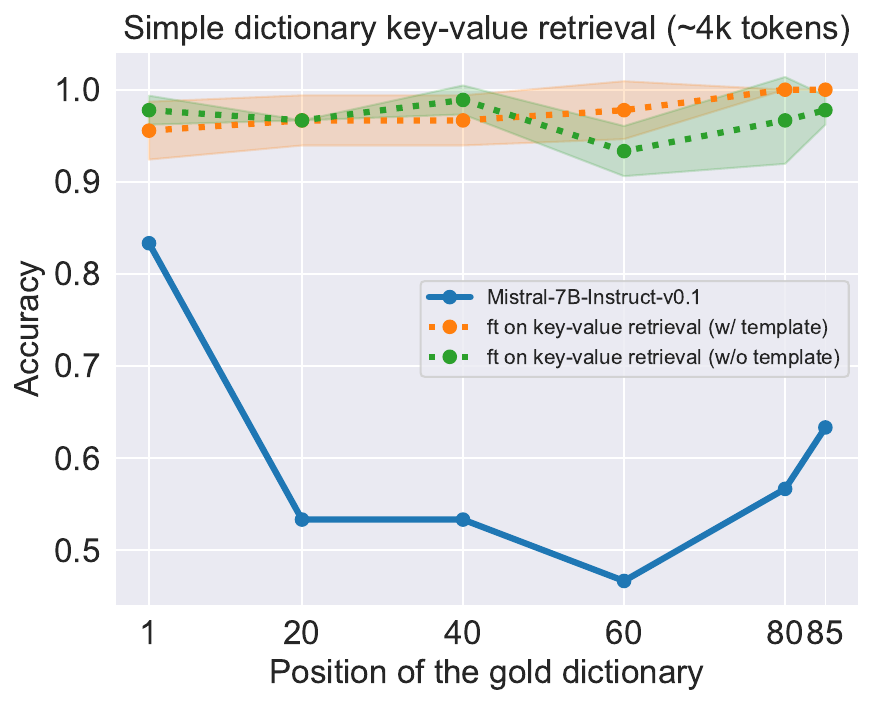}
    \caption{Mistral 7B and the finetuned versions on simple dictionary key-value retrieval.}
    \label{fig:mistral-simpledict}
\end{figure}
For Mistral 7B, we choose simple dictionary key-value retrieval as the task to finetune on. We use two prompting strategies to prepare the dataset: with and without an answer template as described in Section \ref{sec:synthetic_data}. For each prompting strategy we generate $3$ different datasets using the same configuration but with different seeds. Each dataset consists of $350$ simple dictionary key-value retrieval tasks (roughly 4K tokens in each task). Each task has $85$ dictionaries and each dictionary has $3$ to $4$ keys. Each key and value is an integer of $3$ to $4$ digits. We finetune Mistral 7B on all attention layers and use a global batch size of $16$ and finetune the model for $2$ epochs on each dataset with learning rate $5\times10^{-6}$. For evaluation results, we average across $3$ runs, each with different training data and seed. \par
For GPT-3.5 Turbo, we choose multi-subkey key-value retrieval as the task to finetune on. For each prompting strategy, we generate $2$ different datasets. Each dataset consists of $150$ multi-subkey key-value retrieval tasks (roughly 4K tokens in each task). Each task has $49$ dictionaries. We finetune GPT-3.5 Turbo for $2$ epochs on each dataset using OpenAI API. For evaluation results, we average across $2$ runs.
\label{sec:mistral_ft}

\end{document}